# PFformer: A Position-Free Transformer Variant for Extreme-Adaptive Multivariate Time Series Forecasting

Yanhong Li[1] and David C. Anastasiu[1]

[1]Department of Computer Science and Engineering, Santa Clara University, Santa Clara, CA, USA

**Multivariate time series (MTS) forecasting is vital in fields like weather, energy, and finance. However, despite deep learning advancements, traditional Transformer-based models often diminish the effect of crucial inter-variable relationships by singular token embedding and struggle to effectively capture complex dependencies among variables, especially in datasets with rare or extreme events. These events create significant imbalances and lead to high skewness, complicating accurate prediction efforts. This study introduces PFformer, a position-free Transformer-based model designed for single-target MTS forecasting, specifically for challenging datasets characterized by extreme variability. PFformer integrates two novel embedding strategies: Enhanced Feature-based Embedding (EFE) and Auto-Encoder-based Embedding (AEE). EFE effectively encodes inter-variable dependencies by mapping related sequence subsets to high-dimensional spaces without positional constraints, enhancing the encoder's functionality. PFformer shows superior forecasting accuracy without the traditional limitations of positional encoding in MTS modeling. We evaluated PFformer across four challenging datasets, focusing on two key forecasting scenarios: long sequence prediction for 3 days ahead and rolling predictions every four hours to reflect real-time decision-making processes in water management. PFformer demonstrated remarkable improvements, from 20% to 60%, compared with state-of-the-art models.**

**Deep Learning | Time Series Forecasting | Hydrology | Transformer.**

## Introduction

Effective time series forecasting is critical for strategic decision-making in many industries, including meteorology (1, 2), financial markets (3, 4), and energy management (5). Advancements in deep learning have resulted in a wave of complex multivariate time series (MTS) models that have greatly improved forecasting accuracy by utilizing intricate data relationships, especially where sporadic but major extreme events (6, 7) such as flash floods and droughts exist. After its immense success in language processing and computer vision, the Transformer model (8), known for its strong capabilities in depicting pairwise dependencies in sequences, is now making significant inroads into time series forecasting (9–11). However, these models typically compress multiple variables of the same timestamp into a single token, potentially obscuring vital multivariate correlations. As highlighted by Liu et al. (12) and Zhang and Yan (13), data points within the same timestamp often represent distinct physical phenomena and are measured inconsistently. This embedding into a singular token can erase crucial inter-variable relationships, and the localized receptive field of a single timestamp may not capture beneficial information effectively, especially when dealing with events that are not aligned in time. Moreover, the application of permutation-invariant attention mechanisms, which fail to consider the importance of sequence order, may not be suitable for analyzing temporal data where order significantly impacts series variations. Consequently, these flaws can limit the Transformer's capacity to represent and generalize across various multivariate time series effectively. Even with closely connected multivariate inputs available, like the streamflow and rain time series, the Transformer struggles to use these correlations to accurately predict time series with extreme events (14).

In response to these challenges, we propose PFformer, a position-free Transformer variant for adaptive multivariate time series forecasting. The PFformer model leverages a position-free embedding strategy that allows for a more flexible and precise representation of both time and variable dependencies. By abstracting away from fixed positional encodings, the model can better distinguish and integrate relevant features across both dimensions, enhancing its predictive capabilities. Our approach is designed to not only preserve but also accentuate the inherent dependencies within the data, enabling the PFformer to deliver superior forecasting performance on highly skewed datasets. The main contributions of this work are as follows:

- By employing EFE and AEE, PFformer overcomes traditional challenges associated with positional encoding and enables a more flexible and effective handling of spatial-temporal relationships.

- The PFformer model incorporates a novel clustering-based importance enhanced sampling strategy that adeptly pinpoints critical features and trends within datasets by relying on the learned mixture distribution of the data.

- To improve the model's robustness to severe events, PFformer innovatively uses AEE to emphasize short-

term predictions in the loss penalty, which makes the auxiliary variables more accountable for the overall accuracy of the model.

We evaluated PFformer on four separate datasets and found that PFformer significantly outperforms state-of-the-art baselines by 20% to 60%. Additionally, we carried out several ablation studies to understand the effects of specific design decisions.

## Related work

To fully leverage auxiliary variables in prediction, multivariate time series studies have utilized various algorithms, from traditional methods like vector autoregression and multivariate exponential smoothing to advanced deep learning methods. Vector autoregressive (VAR) models (15) are statistical models that assume linear dependencies both across different dimensions and over time. Wang et al. (16) developed a hybrid model that combines Empirical Mode Decomposition (EMD), Ensemble EMD (EEMD), and ARIMA for long-term streamflow forecasting. Additionally, graph neural networks (GNNs) (17) explicitly capture cross-dimensional dependencies by combining temporal and graph convolutional layers. Recently, transformer-based techniques such as Autoformer (18) and Reformer (19) have been proposed for long-term forecasting, offering sophisticated dependency discovery and modeling capabilities. Informer (9) introduced a ProbSparse self-attention mechanism and a generative-style decoder to significantly increase inference speed for long-sequence predictions. FEDFormer (10) improved long-term forecasting by randomly selecting a fixed number of Fourier components to capture the global characteristics of time series. However, recent research suggests that simpler linear models (11, 20) may outperform these complex approaches. Other methods, including representation learning (14) and hybrid techniques (21–23), have also been explored for long-term forecasting. PatchTST (24) enhances time series modeling by using patching techniques to extract local semantics and maintain channel independence. Crossformer (13) features a cross-scale embedding layer and Long Short Distance Attention (LSDA) to effectively capture cross-time and cross-variable dependencies in MTS, and iTransformer (12) optimizes Transformer inputs to improve time-series modeling, focusing on more accurate data interpretation and prediction. Despite extensive research on time series prediction, deep learning models face difficulties when dealing with time series data that contain rare or extreme occurrences because of the obvious imbalance in the dataset. Predicting these types of data is notably difficult as their distribution is heavily influenced by extreme values, leading to high skewness. This calls for the creation of specialist models intended for precise forecasting of extreme events. An and Cho (25) proposed a novel method that focused on anomaly detection using reconstruction probability as a lens. This approach cleverly takes into account the data distribution's intrinsic variability. Similar to this, the Uber TSF model (26) automatically extracts extra features from the auto-encoder LSTM network, priming it to capture intricate time-series dynamics during large-scale events. New inputs are then fed into the LSTM forecaster for prediction. Ding et al. (27) focused their research on enhancing the deep learning models' ability to identify and forecast exceptional events. They used a technique that modifies predictions according to how closely the current data resembles extreme occurrences that have been observed in the past. Additionally, Additionally, an earlier model we designed, NEC+ (28) is specifically designed to accommodate extreme values in hydrologic flow prediction by employing the Gaussian mixture model (GMM) (29) and training three predictors in parallel. Another model we previously proposed, DAN (14), learns and merges rich representations to adaptively predict streamflow.

However, few of these prior works have addressed both long-term predictions in prolonged sequences and used auxiliary variables effectively to enhance extreme value prediction. To bridge this gap, we propose PFformer to address the challenges of highly skewed single-target MTS forecasting in the hydrology domain. Experiments on four real-life hydrologic streamflow datasets show that PFformer significantly outperforms state-of-the-art methods for hydrologic and long-term time series prediction.

## Preliminaries

**Problem Statement.** Suppose we have a collection of $m$ ($m >= 1$) related univariate time series, with each row in the input matrix corresponding to a different time series. We are going to predict the next $h$ time steps for the first time series $x_1$, given historical data from multiple length-$t$ observed series. The problem can be described as,

$$\begin{bmatrix} x_{1,1} & \cdots & x_{1,t} \\ x_{2,1} & \cdots & x_{2,t} \\ \vdots & \ddots & \vdots \\ x_{m,1} & \cdots & x_{m,t} \end{bmatrix} \in \mathbb{R}^{m \times t} \rightarrow [x_{1,t+1}, \ldots, x_{1,t+h}] \in \mathbb{R}^h,$$

where $x_{i,j}$ denotes the value of time series $i$ at time $j$. The matrix on the left are the inputs, and $x_{1,t+1}$ to $x_{1,t+h}$ are the outputs of our method. We first define this task by modeling the objective time series $x_1$ as the *ordinary series* and the group of related time series $x_2$ to $x_m$ as *auxiliary series*.

**Table 1.** Input Stream Data Statistics

| Statistic / Stream | Ross | Saratoga | UpperPen | SFC |
|---:|---:|---:|---:|---:|
| min | 0.00 | 0.00 | 0.00 | 0.00 |
| max | 1440.00 | 2210.00 | 830.00 | 7200.00 |
| mean | 2.91 | 5.77 | 6.66 | 20.25 |
| std. deviation | 24.43 | 26.66 | 21.28 | 110.03 |
| skewness | 19.84 | 19.50 | 13.42 | 18.05 |
| kurtosis | 523.16 | 697.78 | 262.18 | 555.18 |

**Data Descriptions.** Our study uses a hydrologic dataset first introduced in (14) that captures streamflow from four California streams: Ross, Saratoga, UpperPen, and SFC. Given California's lack of rainfall during the summer, we follow the same problem design as in (14) and focus on forecasting the months from September to May, deliberately excluding the summer period. Data for training and validation was drawn from January 1988 to August 2021 and we aim to accurately project the streamflow for the subsequent year (September 2021 to May 2022), with predictions made every four hours. Each prediction estimates the upcoming 3 days based on the preceding 15 days of data. The performance metrics we employed are Root Mean Square Error (RMSE) and Mean Absolute Percentage Error (MAPE). We add 1 to both the true value and the predicted value to avoid instability in the calculation when the true values are close to zero. Since the sensors measure the streamflow and precipitation every 15 minutes, we are attempting a lengthy forecasting horizon ($h = 288$). Table 1 shows several statistics of our input time series, which provide valuable insights into the shape and distribution of the data, including min, max, mean, median, variance, skewness, and kurtosis. The presence of high skewness and kurtosis values suggests that our data exhibit significant asymmetry and departure from the symmetric bell-shaped curve of a Normal distribution. Specifically, the positive skewness values indicate that the distribution is skewed to the right, resulting in a longer tail on the right side. This implies that there are more extreme values or outliers on the higher end of the distribution.

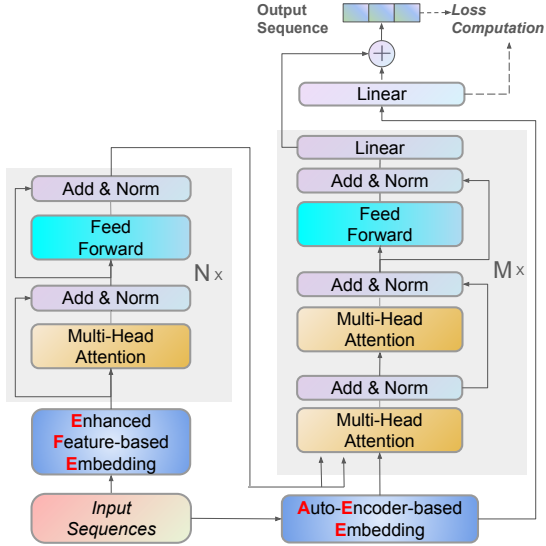

**Fig. 1.** *The PFformer framework*; ⊕ denotes the element-wise addition. The PFformer framework is a transformer-based variant optimized for multivariate time series forecasting. It replaces the encoder's positional encoding layer with Enhanced Feature-based Embedding (EFE) to capture complex inter-variable relationships. In the decoder, Auto-Encoder-based Embedding (AEE) substitutes the positional embedding, enabling direct, fixed-length predictions without error propagation or masking.

## Methods

**PFformer Architecture.** Fig. 1 illustrates the PFformer architecture, a transformer-based framework optimized for multivariate time series forecasting. It consists of two main components: the encoder and the decoder, both structured to enhance forecasting accuracy through advanced embedding techniques.

**Encoder.** As shown in the left part of Fig. 1, the encoder section of our model closely follows the standard Transformer encoder design aside from the embedding layer, which uses Enhanced Feature-based Embedding (EFE). This strategy maps input sequences to high-dimensional spaces without positional encoding and effectively captures complex inter-variable relationships. These are then processed through layers of multi-head attention and feed-forward networks, each featuring an "Add and Norm" step for output integration and normalization, crucial for stabilizing the learning process.

**Decoder.** PFformer bypasses the use of target sequences as inputs to avoid error propagation common with recursive prediction methods. Instead, it predicts all outcomes directly, eliminating the need for both masking mechanisms and padding due to the fixed prediction length. The AEE layer is used to create rich representations that enhance input data for the attention mechanisms. Additionally, we simplified the cross-attention layer by applying attention computing directly on the output of AEE in the transformer decoder, enabling effective learning from the richly represented embedding layers.

**Output.** To enhance the model's adaptability to extreme values, the AEE module also plays a critical role in the loss calculation. Specifically, the output from AEE passes through a linear layer, which is combined with the output from the decoder's linear layer to form the final output, as further described in the Multiple-Objective Loss Function section.

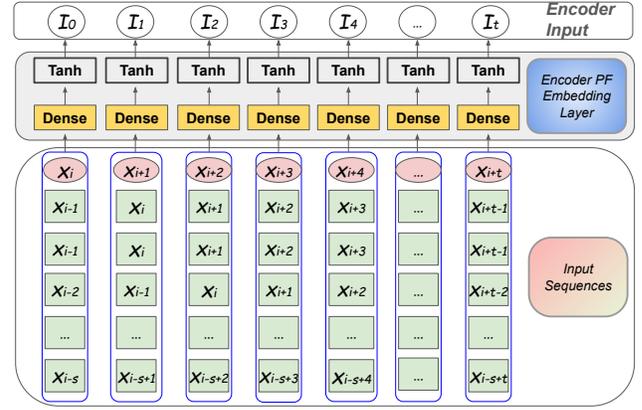

**Fig. 2.** *PFformer encoder embedding module.* $X$ and $A$ are the stream and rain value sequences in our dataset. The output of EFE is the input to the transformer encoder.

**Enhanced Feature-Based Embedding (EFE).** Fig. 2 shows the EFE module in PFformer. For an input sequence of length $t$, taking the first time point as an example, EFE combines the predicted sequence value at this time point, corresponding values from auxiliary sequences, and $s$ preceding values from each of these auxiliary sequences. These elements are concatenated to form a multi-spatial subsequence unique to that time point. This subsequence is then processed through a dense layer coupled with a nonlinear activation function. Then, the output of the EFE module serves as the input vector for the encoder at this fixed time point. Across all time points, this forms a high-dimensional input sequence for the encoder layer. The EFE mechanism focuses on cap-turing the intricate relationships between different time series variables across time. By constructing subsets of related sequences from preceding time points, EFE directly encapsulates inter-variable dependencies within the embeddings for each time instance. Such an approach is particularly advantageous for the Transformer encoder, which can then operate on these self-contained embeddings without the usual constraints imposed by positional encoding.

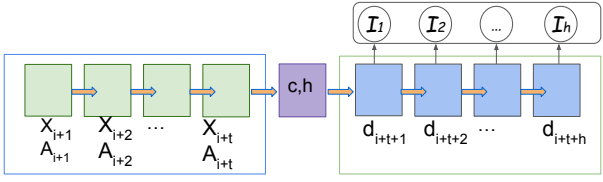

**Fig. 3.** *AEE module.* The input of the AEE encoder is the aligned multivariate series. $X$ and $A$ are the stream and rain value sequences in our dataset. The input of the AEE decoder is the time stamp of the forecasted series.

**Auto-Encoder-Based Embedding (AEE).** Fig. 3 illustrates the Auto-Encoder-Based Embedding (AEE) layer, which is designed to handle aligned multivariate time series data, such as streamflow and rainfall in our dataset. The encoder module takes as input the aligned multivariate series, where $X$ represents the streamflow sequence, and $A$ corresponds to the rainfall sequence. These sequences are processed step-by-step to generate the latent states $c$ and $h$, which summarize the temporal patterns and dependencies in the input data. The decoder module then uses the latent states, along with the time stamps of the forecasted series, to predict future values.

The output of the AEE fulfills two key roles. First, it is integrated with the outputs from the PFformer decoder, enhancing the overall prediction accuracy. Second, its capacity for short-term prediction is measured as a component of the loss function. This inclusion acts as a penalty term, underscoring the critical role of auxiliary variables in refining short-term forecast accuracy.

**Clustering-Based Oversampling Policy.** Since the total length of each time series in our dataset is approximately 1.4 million, the sampling strategy is crucial during model training. However, simply oversampling extreme values can degrade overall prediction quality for the rest of the time series, as previously demonstrated in research. To address this, we propose a Clustering-Based Oversampling Policy which aims to capture significant data points based on statistic distributions. We employ a Gaussian Mixture Model (GMM) to cluster the data into $M$ clusters with mean values $\mu_1, \mu_2, \ldots, \mu_M$. Since the cluster with the highest mean value $z$ represents those extreme values, we marked a data point as important if its value exceeds $\eta \times z$. For each peak point, we treat step size $s$ and scope $\nu$ as hyperparameters. The sampling starts $\nu/2$ points to the left of an identified peak, and samples are collected every $s$ points, effectively generating $\nu/s$ samples around each peak.

To manage the volume of oversampled data, we set a ratio, $os$, which caps the oversampled data at $os\%$ of the total training set volume. We employ grid search to determine the optimal value of $os$ to ensure our sampling method remains targeted and efficient without overwhelming the model with excessive data.

**Multiple-Objective Loss Function.** To force the AEE to learn rich representations, we use multiple loss functions (30–32) when training the PFformer model. We build our loss items as follows,

$$\mathcal{L}_1 = RMSE(\hat{y}_{aux}[:s], y[:s]), \quad \mathcal{L}_2 = RMSE(\hat{y}, y).$$

where $\hat{y}$ is the output of PFformer, $\hat{y}_{aux}$ is the output of the AEE module after a linear transformation, and $s$ represents the length of the short-term interval, which is set to 16 (4 hours) in our experiments. The $\mathcal{L}_1$ regularization component is responsible for short-term prediction accuracy, while $\mathcal{L}_2$ focuses on the overall model accuracy. Then, the overall loss is composed as,

$$\mathcal{L} = \lambda \times \mathcal{L}_1 + \mathcal{L}_2, \qquad \lambda = max(-1 \cdot e^{\frac{epoch}{45}} + \alpha, \beta).$$

$\lambda$ acts as a scaling factor for regularization, initially set high to steer the AEE to be more responsible for short-term prediction accuracy. This 'teacher mode' diminishes over time; $\lambda$ starts at $\alpha$ and decreases to $\beta$ over epochs.

## Evaluation

**Experimental Settings.** This study utilized the same time series data as in (14). In our experiments, we consistently set $N = 3$, $M = 1$, and $\eta = 1.2$ across all datasets (Ross, Saratoga, UpperPen, and SFC). The oversampling percentages $os\%$ were adjusted to 20% for Ross, Saratoga, and UpperPen, and to 15% for SFC, reflecting the unique distributions of each dataset. The oversampling policies, denoted as $(s, v)$, and the $\alpha$ and $\beta$ values for the regularization parameters were tailored to the distinct characteristics of each dataset. We performed a grid search with $s \in \{1, 2, 4\}$, $v \in \{4, 8, 16\}$, $\alpha \in \{1, 1.2, 1.5, 1.8, 2, 2.2\}$, and $\beta \in \{0.5, 0.6, \ldots, 1\}$ and obtained the best results at $(s, v, \alpha, \beta)$ of (1,8,1.5,0.9), (2,16,1.8,0.8), (4,16,2.0,0.8), and (2,16,2.0,0.9) for the Ross, Saratoga, UpperPen, and SFC datasets, respectively. The hidden dimensions for the attention and linear layers were set to [384, 268, 288, 300], and for the AEE LSTM layer to [384, 268, 320, 256]. Furthermore, the AEE featured one LSTM layer for SFC and two for the other sensors. All models were trained for a maximum of 40 epochs with early stopping triggered after four consecutive epochs without improvement. We used the Adam gradient descent algorithm with a starting learning rate of 0.0005, decaying by 0.9 after every epoch, and set the batch size to 24 for SFC, 48 for Saratoga and UpperPen, and 96 for Ross. All models were trained using PyTorch 1.11.0+cu102 on a Supermicro SYS-420GP-TNAR+ system equipped with NVIDIA HGX A100 8-way GPUs (80 GB RAM each) running Rocky Linux 9.4 (Blue Onyx), but only used one GPU for all training and inference. We also verified training and inference of our model on a system with one NVIDIA V100 GPU (32 GB RAM).

Table 2. 3-day/4-hour Long-Term ($h = 288$) Series Forecasting Results

| Methods | RMSE | | | | | | | | MAPE | | | | | | | |
|---|---|---|---|---|---|---|---|---|---|---|---|---|---|---|---|---|
| | Ross | | Saratoga | | UpperPen | | SFC | | Ross | | Saratoga | | UpperPen | | SFC | |
| | 3 d | 4 h | 3 d | 4 h | 3 d | 4 h | 3 d | 4 h | 3 d | 4 h | 3 d | 4 h | 3 d | 4 h | 3 d | 4 h |
| FEDformer | 6.01 | 3.95 | 6.01 | 4.82 | 3.05 | 2.55 | 23.54 | 17.11 | 2.10 | 2.05 | 1.55 | 1.54 | 1.87 | 1.75 | 2.35 | 2.16 |
| Informer | 7.84 | 6.76 | 5.04 | 3.78 | 5.88 | 5.00 | 39.89 | 23.21 | 4.05 | 4.71 | 1.43 | 1.54 | 4.10 | 3.99 | 8.64 | 3.61 |
| Nlinear | 6.10 | 2.76 | 5.23 | 4.13 | 1.57 | 0.51 | 18.47 | 5.08 | 1.99 | 0.52 | 0.83 | 0.82 | 0.45 | 0.16 | 0.92 | 0.52 |
| Dlinear | 7.16 | 3.31 | 4.33 | 1.79 | 3.53 | 1.35 | 21.62 | 8.75 | 3.10 | 1.15 | 1.40 | 0.65 | 2.35 | 0.69 | 2.74 | 1.45 |
| LSTM-Atten | 7.35 | 6.84 | 6.49 | 5.59 | 6.35 | 4.75 | 34.17 | 23.09 | 3.74 | 4.10 | 1.80 | 1.79 | 4.76 | 3.67 | 9.90 | 6.25 |
| NEC+ | 9.44 | <u>2.07</u> | 1.88 | <u>0.26</u> | 2.22 | <u>0.33</u> | 17.00 | **2.36** | 4.80 | <u>0.45</u> | 0.17 | <u>0.07</u> | 0.95 | <u>0.06</u> | 1.07 | <u>0.07</u> |
| iTransformer | 4.56 | 2.14 | 2.37 | 0.94 | 1.12 | 0.58 | 17.04 | 11.00 | 0.57 | 0.43 | 0.27 | 0.18 | <u>0.11</u> | <u>0.06</u> | 0.47 | 0.54 |
| DAN | <u>4.25</u> | 2.61 | <u>1.80</u> | 0.62 | <u>1.10</u> | 0.43 | <u>15.23</u> | 3.73 | **0.07** | 0.46 | <u>0.14</u> | 0.22 | 0.15 | 0.07 | <u>0.26</u> | 0.22 |
| PFformer | **4.21** | **1.52** | **1.69** | **0.22** | **1.01** | **0.24** | **14.98** | <u>2.86</u> | <u>0.10</u> | **0.03** | **0.10** | **0.04** | **0.06** | **0.01** | **0.18** | **0.06** |

**Baseline Methods.** We compared our proposed method, PFformer[1], against a wide array of state-of-the-art time series and hydrologic prediction methods we discussed in the Related Works section. FEDFormer (10) enhances the standard Transformer by incorporating seasonal-trend decomposition. Informer (9) introduces a prob-sparse self-attention mechanism tailored for long-term time series prediction. NEC+ (28) utilizes LSTM-based models optimized for hydrologic time series prediction in series with extreme events. NLinear (11) is an effective linear model with one order difference preprocessing for long-term time series, while DLinear (11) focuses on decomposing trends for similar applications. Attention-LSTM (33) serves as a state-of-the-art multivariate model in hydrology. Finally, iTransformer (12) achieved state-of-the-art results on challenging multivariate time series prediction challenges, and DAN (14) learns and merges rich representations to adaptively predict streamflow.

**Main Results.** The experimental results, detailed in Table 2, highlight the best and second-best values for each metric, marked in bold and underline, respectively. Overall, PFformer consistently outperforms other models. While models tailored for hydrologic data forecasting, like NEC+ and DAN, perform better than more generic time series prediction models, they do not match the effectiveness of PFformer. Specifically, PFformer not only surpasses DAN across all RMSE metrics but also achieves remarkable improvements of 23% to 64% in 4-hour rolling prediction RMSE. Compared to NEC+ and iTransformer, PFformer shows an average increase of 22% and 37% across all RMSE metrics respectively, showing its extreme-adaptive ability.

Transformer-based methods such as FEDFormer and Informer struggle significantly with datasets that exhibit large variance and fail to adapt to extreme values effectively, not fully leveraging the attention mechanism's potential. While iTransformer effectively utilizes variate tokens to capture multivariate correlations through attention, it risks losing intra-variable temporal relationships by focusing solely on inverted dimensions. This limitation is significant in forecasting time series with extreme values, where both inter-variable relationships and intra-variable temporal sequences are crucial for accuracy. Our EFE and AEE address this by preserving both. In contrast, fully connected networks like NLinear and DLinear, which decompose data into main trends and residuals, perform better than some Transformer-based methods, particularly in rolling prediction scenarios. However, they still fall short of achieving high accuracy.

**3-Days Prediction.** In the realm of single-shot 3-day predictions, PFformer consistently outperforms baselines in RMSE across all four sensors. Notably, when compared to DAN, PFformer achieves a 60% improvement in MAPE for the UpperPen dataset. Furthermore, against NEC+, PFformer demonstrates substantial enhancements on the Ross and UpperPen datasets, with over 50% improvement in RMSE and over 90% in MAPE. For the Saratoga dataset, PFformer outperforms iTransformer by 28% and 62% in RMSE and MAPE, respectively.

**4-Hour Rolling Prediction.** In the scenario of 4-hour rolling predictions, which is more common in practical applications, the PFformer model showcases exceptional performance. Compared to DAN, it achieves an average RMSE improvement of 43% and a MAPE improvement of 83%. Against NEC+, PFformer demonstrates a 22% improvement in RMSE and a 55% increase in MAPE. Compared with iTransformer, PFformer is superior by 59% and 85% in RMSE and MAPE, respectively. This scenario holds significant practical value, reflecting the model's strong applicability in real-world settings.

**Visual Analysis.** Fig. 4 shows some example predictions for PFformer and the next two best performing algorithms. As seen in the figure, PFformer's predictions closely match the ground truth, particularly in datasets with considerable oscillations. Notably, PFformer more accurately captures short-term oscillations and severe values than DAN, as evidenced by lower RMSE values. This improved performance is primarily due to the rich expressive capabilities of the embeddings mixed with the attention mechanism.

## Ablation Study

In this ablation study, we aim to explore several key research questions that address the prediction effectiveness of our model, including (1) the impact of our clustering-based oversampling policy on model performance; (2) the significance of subsequence length $s$ in our model's EFE module; (3) the influence of loss function parameters on its predictive

---

[1]Code for our method is at https://github.com/davidanastasiu/pfformer.

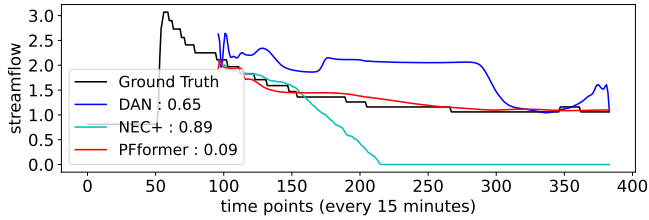 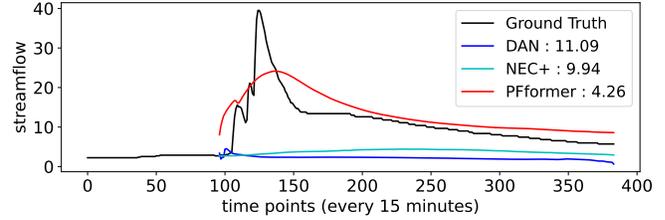

**Fig. 4.** Comparative examples with the best baselines reveal that PFformer excels in predicting streamflow values with greater accuracy, both in the short and long terms.

accuracy; and (4) the overall effect of both the EFE and AEE modules.

**Effect of Parameters.** To address the first three research questions, we first established the optimal parameter combination for the Ross dataset as our baseline. Then, fixing all other parameters, adjusted only one parameter at a time to analyze its influence on model performance. Fig. 5 shows our results for oversampling percentages $os \in \{10, 20, 30, 40, 50, 60\}$ (left), the EFE susequence length $s \in \{20, 40, 60, 80, 100, 120\}$ (center), and the loss multiplier parameter $\alpha \in \{0, 1, 1.5, 2.0, 2.5, 3.0\}$ (right), for both 3-day prediction (y1-axis) and 4-hour prediction (y2-axis) scenarios.

**Oversampling.** For the 3-day RMSE, the model performs best at a 20% rate, while short-term predictions are more sensitive to extreme values in the training set, but a 20% rate is the best overall.

**EFE Subsequence Length.** Regarding the EFE parameter $s$, theoretically, a larger $s$ contains more information. By adjusting the accompanying hidden size and number of layers, a better model might be found, but this also increases redundancy and computation load. In our experiments, $s = 60$ was best.

**Loss Regularization.** Higher values of the loss regularization parameter $\alpha$ force PFformer to focus more on the accuracy of short-term predictions, but should be balanced with the need for overall good long-term predictions. In our experiment, $\alpha = 1.5$ produced the best results in both 3-day ahead and 4-hour predictions.

**Table 3.** PFformer $RMSE$ With Different Embeddings

| Dataset | Type | EFE/AEE | Position | Token |
|---|---|---|---|---|
| Ross | 3-day | 4.21 | 4.41 | 4.55 |
|  | rolling | 1.52 | 2.24 | 2.61 |
| SFC | 3-day | 14.98 | 21.68 | 21.38 |
|  | rolling | 2.86 | 17.13 | 16.72 |
| UpperPen | 3-day | 1.01 | 3.91 | 2.44 |
|  | rolling | 0.24 | 3.92 | 1.47 |
| Saratoga | 3-day | 1.69 | 2.59 | 3.88 |
|  | rolling | 0.29 | 1.71 | 3.88 |

**Effect of Architecture.** In Table 3, the first column shows the RMSE values obtained when using both the EFE and AEE layers. In the second column, we removed EFE and AEE and reverted to using the original Transformer's combination of position embedding and token embedding. In the third column, we eliminated position embedding altogether and solely utilized token embedding. The results show that the EFE and AEE introduced by PFformer significantly enhance overall predictive performance by enabling the attention layer to focus on short-term performance in a rolling prediction mode. Moreover, the position embedding inherent in traditional Transformers has an inconsistent effect on time series forecasting, failing to enhance performance on datasets like SFC and UpperPen.

## Conclusion

This study has successfully demonstrated the efficacy of position-free trainable embedding techniques—Enhanced Feature-based Embedding (EFE) and Auto-Encoder-based Embedding (AEE)—in improving hydrologic flow prediction. By comparing these techniques against traditional Transformer embeddings and other state-of-the-art methods, we have shown that our method significantly enhances forecasting accuracy, especially in scenarios with extreme values. Our findings suggest that traditional positional information may be less crucial for time series than previously thought, indicating a promising direction for future research to further refine these models for broader applications in complex time series forecasting scenarios.

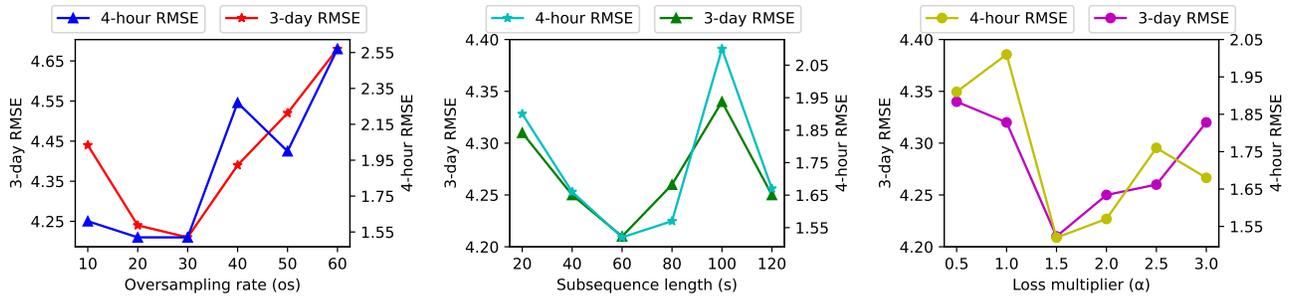

**Fig. 5.** PFformer RMSE on Ross given different oversampling percentage $os$ (left), EFE subsequence length $s$ (center), and loss multiplier $\alpha$ (right) values, under both 3-day prediction (y1-axis) and 4-hour prediction (y2-axis) scenarios.